\documentclass[pmlr]{jmlr}

\usepackage{rotating}
\usepackage{longtable}
 \usepackage{booktabs}
 
\usepackage[load-configurations=version-1]{siunitx} 

\makeatletter
\def\set@curr@file#1{\def\@curr@file{#1}} 
\makeatother

\usepackage{lipsum}
\usepackage{xcolor}
\usepackage{mathtools}

\usepackage{soul}
\sethlcolor{black}
\makeatletter
\DeclareMathOperator*{\argmax}{arg\,max}

\theorembodyfont{\upshape}
\theoremheaderfont{\scshape}
\theorempostheader{:}
\theoremsep{\newline}

\jmlrvolume{}
\jmlryear{2023}
\jmlrworkshop{}

\title[Deep Attention Q-Network for Personalized Treatment Recommendation]{Deep Attention Q-Network for Personalized Treatment Recommendation}

\author{\Name{Simin Ma}
       \Email{sma318@gatech.edu}\\ 
       \Name{Junghwan Lee}
       \Email{jlee3541@gatech.edu}\\ 
       \Name{Nicoleta Serban}
       \Email{nicoleta.serban@isye.gatech.edu}\\ 
       \Name{Shihao Yang}
       \Email{shihao.yang@isye.gatech.edu}\\ 
       \addr H.Milton School of Industrial and Systems Engineering\\
       Georgia Institute of Technology\\
       }

\begin{document}

\maketitle

\begin{abstract}
Tailoring treatment for individual patients is crucial yet challenging in order to achieve optimal healthcare outcomes. Recent advances in reinforcement learning offer promising personalized treatment recommendations; however, they rely solely on current patient observations (vital signs, demographics) as the patient's state, which may not accurately represent the true health status of the patient. This limitation hampers policy learning and evaluation, ultimately limiting treatment effectiveness. In this study, we propose the Deep Attention Q-Network for personalized treatment recommendations, utilizing the Transformer architecture within a deep reinforcement learning framework to efficiently incorporate all past patient observations. We evaluated the model on real-world sepsis and acute hypotension cohorts, demonstrating its superiority to state-of-the-art models. The source code for our model is available at \url{https://github.com/stevenmsm/RL-ICU-DAQN}.
\end{abstract}

\section{Introduction}
Intensive Care Unit (ICU) treatment recommendation is a critical task, as it plays a vital role in the management and care of critically ill patients. Current state-of-the-art methods for providing ICU treatment recommendations primarily involve rule-based protocols and evidence-based clinical guidelines, which are informed by randomized controlled trials (RCTs), systematic reviews, and meta-analyses. However, RCTs may not be available or definitive for many ICU conditions \citep{nemati2016optimal}, and individual patients may respond differently to the same treatment strategy \citep{laffey2018negative}. Therefore, more personalized and effective treatment plans that take into account the dynamic nature of patients' conditions and the potential presence of multiple comorbidities are needed in the ICU setting to benefit critically ill patients. Recent developments in artificial intelligence (AI) have demonstrated various successful applications in the healthcare domain, such as diagnosis \citep{burke2019feasibility, laserson2018textray}, treatment \citep{choi2016doctor, fan2019automatic}, and resource management \citep{watanabe2019improved}. Reinforcement learning (RL), in particular, is well-suited for learning optimal individual treatment interventions. RL involves sequential decision-making in an environment with evaluative feedback, with the goal of maximizing an expected reward \citep{montague1999reinforcement}. RL shares the same goal as clinicians: making therapeutic decisions to maximize a patient’s probability of a good outcome. Therefore, RL has many desirable properties and has already shown its success in providing sequential treatment suggestions in various ICU settings, such as optimal dosing of medication \citep{komorowski2018artificial, raghu2017continuous, peng2018improving, raghu2017deep, huang2022reinforcement, padmanabhan2015closed, borera2011adaptive, nemati2016optimal}, optimal timing of intervention \citep{prasad2017reinforcement, yu2019inverse}, optimal choice of medication \citep{wang2018supervised}, and optimal individual target lab value \citep{weng2017representation}, among others. The findings from these studies all suggest that if physicians followed the RL policy, the estimated hospital mortality could be improved.

However, the patient-clinician interactions in the aforementioned studies are all modeled as Markov decision processes (MDPs), while in practice, the pathology is often complex, and the "true" underlying states of the patients are latent and can only be observed through emitted signals (observations) with some uncertainty. The challenge is that the ICU setting might not be a fully observable environment for RL agents; this could be due to a variety of factors such as noisy measurements, omission of relevant factors, and the incongruity of the frequencies and time-lags among the considered measurements \citep{li2019optimizing}. To alleviate the issue of a partially observable environment, RL agents may need to remember (some or possibly all) previous observations. As a result, RL methods typically add some sort of memory component, allowing them to store or refer back to recent observations to make more informed decisions. For example, recurrent neural networks (RNNs) have been used to encode histories \citep{wierstra2007solving, hausknecht2015deep, peng2018improving} or belief transitions \citep{igl2018deep, li2019optimizing}. However, this creates further issues: RNNs can be subject to gradient exploding/vanishing and can be difficult to train. Recent advancements in natural language processing have led to the development of RL studies that employ the powerful Transformer architecture \citep{vaswani2017attention}. For instance, \citet{esslinger2022deep} takes a step further from \citet{hausknecht2015deep}'s deep RL approach by replacing the RNNs with Transformer. Similarly, \cite{chen2021decision} abstracts the RL problem as a sequential modeling problem, and solve it by a variant of Transformer architecture. While these methods have demonstrated improved performance, they  lack interpretability between the states and actions, which is a crucial factor in the healthcare settings. This interpretability issue arises due to the black-box nature of the models, which makes it challenging to understand how the decisions are made. Thus, it is important to develop interpretable models that can provide insights into the reasoning behind the decision-making process in healthcare settings.

In this study, we propose a novel data-driven reinforcement learning approach capable of dynamically suggesting optimal personalized clinical treatments for ICU patients. By efficiently memorizing past patient states and actions, our proposed deep reinforcement learning method can identify suitable upcoming actions and interpret the importance of relationships between actions and past observations. Compared to generic approaches for RL \citep{esslinger2022deep, chen2021decision}, the proposed method's Transformer structure is tailor-made for healthcare data, offering improved interpretability. We demonstrate the robustness and efficiency of our proposed method on two ICU patient disease cohorts, sepsis and acute hypotension, and compare the performances against simpler and alternative benchmark approaches. The evaluation results illustrate that our proposed algorithm's learned optimal policy is able to outperform competing policies, with the help of the attention mechanism. Additionally, we observe that the optimal policy can focus on different past observations by visualizing the attention mechanism, further providing interpretability in our proposed approach.

\subsection*{Generalizable Insights about Machine Learning in the Context of Healthcare}

\begin{itemize}
    \item \textbf{Enhancing the resolution of the patient's state space results in more comprehensive and valuable recommendations, ultimately leading to improved treatment effectiveness.} In ICU settings, patient's ``true'' underlying health status might depend more than the current patient's observations, due to various factors. Therefore, it is important to improve the granularity of patient's state representation, when making treatment recommendations that depends on ``true'' patient underlying health status. Our proposed RL approach ``enrich'' the patient state space by letting RL agent efficiently memorize patients' current and past observations and actions in order to provide more effective and robust recommendations.
    \item \textbf{Interpretability in Machine Learning solutions to healthcare is important and challanging.} Machine learning solutions are often lack explanability and interpretability in clinical sense, which hinders clinical insight and subsequent deployment in real-world settings. Our proposed approach is able to focus on past observations (and actions) that indicate worse patient health when making treatment decisions, which aligns with clinician's diagnosis process and makes RL-based treatment search closer to real-world deployment.
\end{itemize}

\section{Related Work}

\subsection{Reinforcement learning for personalized treatment recommendation}
Reinforcement learning (RL) has garnered considerable attention in determining optimal dosages and treatments for patients in the intensive care unit (ICU). Various studies have investigated its application to different medical scenarios, including propofol dosing for surgical patients \citep{borera2011adaptive, padmanabhan2015closed, yu2019inverse}, heparin dosing for patients with cardiovascular diseases \citep{nemati2016optimal, ghassemi2018personalized, lin2018deep}, intravenous (IV) fluids and vasopressors dosing for sepsis patients \citep{komorowski2018artificial, raghu2017deep, raghu2017continuous, raghu2018model, peng2018improving, huang2022reinforcement}, and morphine dosing for patients with at least one pain intensity score \citep{lopez2019deep}. Sepsis, a leading cause of hospital deaths, is a disease that is costly to treat \citep{cohen2015sepsis}. In addition to antibiotics and source control, the use of IV fluids to correct hypovolemia and vasopressors to counteract sepsis-induced vasodilation presents significant challenges. \cite{komorowski2018artificial} initially formulated the personalized optimal dosage for IV fluids and vasopressors as an RL problem with the goal of improving patients' outcomes, solving it with discrete state and action-based value iteration. \cite{raghu2017deep} extended the model to continuous state space and discrete action and proposed solving the problem using Deep Q-learning \citep{mnih2015human}. Subsequent research studies proposed various re-formulations of the problem or explored different RL algorithms, including model-based RL algorithms \citep{raghu2018model}, a combination of model-free deep RL approach and model-based kernel RL approach \citep{peng2018improving}, and extension to continuous action space solved via policy-gradient algorithms \citep{huang2022reinforcement}. To the best of our knowledge, our study is the first to consider modeling patient-clinician interactions as a partially observable environment. Inspired by the Transformer architecture \citep{vaswani2017attention}, we represent all prior patient observations and actions as the patient's state space in our proposed deep RL approach.

\subsection{Background: Deep Q-Learning}
Reinforcement Learning is concerned with learning control policies for agents interacting with unknown environments. Such environments are often formalized as a Markov Decision Processes (MDPs), described by a 4-tuple $(\mathcal{S}, \mathcal{A}, \mathcal{P}, \mathcal{R})$. At each timestep $t$, an agent interacting with the MDP observes a state $s_t\in \mathcal{S}$, and chooses an action at $a_t\in \mathcal{A}$ which determines the reward $r_t \sim \mathcal{R}(s_t, a_t)$ (reward distribution) and next state $s_{t+1}\sim \mathcal{P}(s_t, a_t)$ (state transition probability distribution). The goal of the agent is to maximize the expected discounted cumulative reward, $\mathbb{E}[\sum_t\gamma^tr_t]$, for some discount factor $\gamma\in [0,1)$.

Q-Learning \citep{watkins1992q} is a model-free off-policy algorithm for estimating the long-term expected return of executing an action from a given state in an MDP. These estimated returns are known as Q-values. A higher Q-value indicates an action $a$ is judged to yield better long-term results in a state $s$. Q-values are learned iteratively by updating the current Q-value estimate towards the observed reward plus the max Q-value over all actions $a'$ in the resulting state $s'$:
\begin{equation}
Q(s,a)\coloneqq Q(s,a)+\alpha\left(r+\gamma\max_{a'\in\mathcal{A}}Q(s',a') - Q(s,a)\right)
\label{eq:Q_update}
\end{equation}
In more challenging domains, however, the state-action space of the environment is often too large to be able to learn an exact Q-value for each state-action pair. Instead of learning a tabular Q-function, Deep Q-Networks  (DQN) \citep{mnih2015human} learns an approximate Q-function featuring strong generalization capabilities over similar states and actions, with the help of neural networks. DQN is trained to minimize the Mean Squared Bellman Error:
\begin{equation}
L(\theta) = \mathbb{E}_{s,a,r,s'}\Bigg[\left(r+\gamma\max_{a'\in\mathcal{A}}Q(s',a';\theta') - Q(s,a;\theta)\right)^2\Bigg]
\label{eq:DQN_loss}
\end{equation}
where transition tuples of states, actions, rewards, and future states $(s, a, r, s')$ are sampled uniformly from a replay buffer, D, of past experiences while training. The target $r+\gamma\max_{a'}Q(s',a';\theta')$ invokes DQN’s target network (parameterized by $\theta'$), which lags behind the main network (parameterized by $\theta$) to produce more stable updates.

\section{Methods}
\subsection{Problem formulation}
When an environment does not emit its full state to the agent, the problem can be modeled as a Partially Observable Markov Decision Process (POMDP), described by 6-tuple $(\mathcal{S}, \mathcal{A}, \mathcal{P}, \mathcal{R}, \Omega, \mathcal{O})$. The two additional sets, $\Omega, \mathcal{O}$, represents the observations set and state-observation distributions, respectively. In particular, at each time step $t$, after the agent (in state $s_t\in \mathcal{S}$) interacts with the environment by taking action $a_t\in \mathcal{A}$ and obtain the reward $r_t \sim \mathcal{R}(s_t, a_t)$, the agent moves into the next state $s_{t+1}\sim \mathcal{P}(s_t, a_t)$, but no longer observes true system state and instead receives an observation $o_{t+1} \in \Omega$, generated from the underlying system state $s_{t+1}$ according to the probability distribution $o_{t+1}\sim \mathcal{O}(s_{t+1})$. In ICU setting, $s_{t}$ can be interpreted as the ``true'' underlying patient's health status, while $o_{t}$ is the current observable measurements (vital signs, demographics, etc.). Because agents in POMDPs do not have access to the environment’s full state information, they must rely on the observations $o_t\in\Omega$. In this case, DQN may not learn a good policy by simply estimating the Q-values from the current patient's observation, $o_t$, since it may not be representative enough for the ``true'' underlying patient's health status, $s_t$. Instead, one often needs to consider some form of all patient's historical observations and actions, for instance $\{(o_0, a_0), (o_1, a_1), \ldots, (o_{t-1}, a_{t-1})\}$, to approximate the true current state, $s_t$. Because the history grows indefinitely as the agent proceeds in a trajectory, efficient ways of encoding the history is needed, for example using an agent's belief \citep{igl2018deep}, using recurrent neural network and its variants \citep{hausknecht2015deep, zeng2019navigation}, etc. Here, we incorporate the recently developed Transformer's attention mechanisms \citep{vaswani2017attention} into the Deep Q-Networks, which is able to incorporate histories into the Q-function and reflect the relative importance between the upcoming actions and the histories.

\subsection{Proposed Method: Deep Attention Q-Network}

\begin{figure}[!h]
\centering
\includegraphics[width=0.8\textwidth]{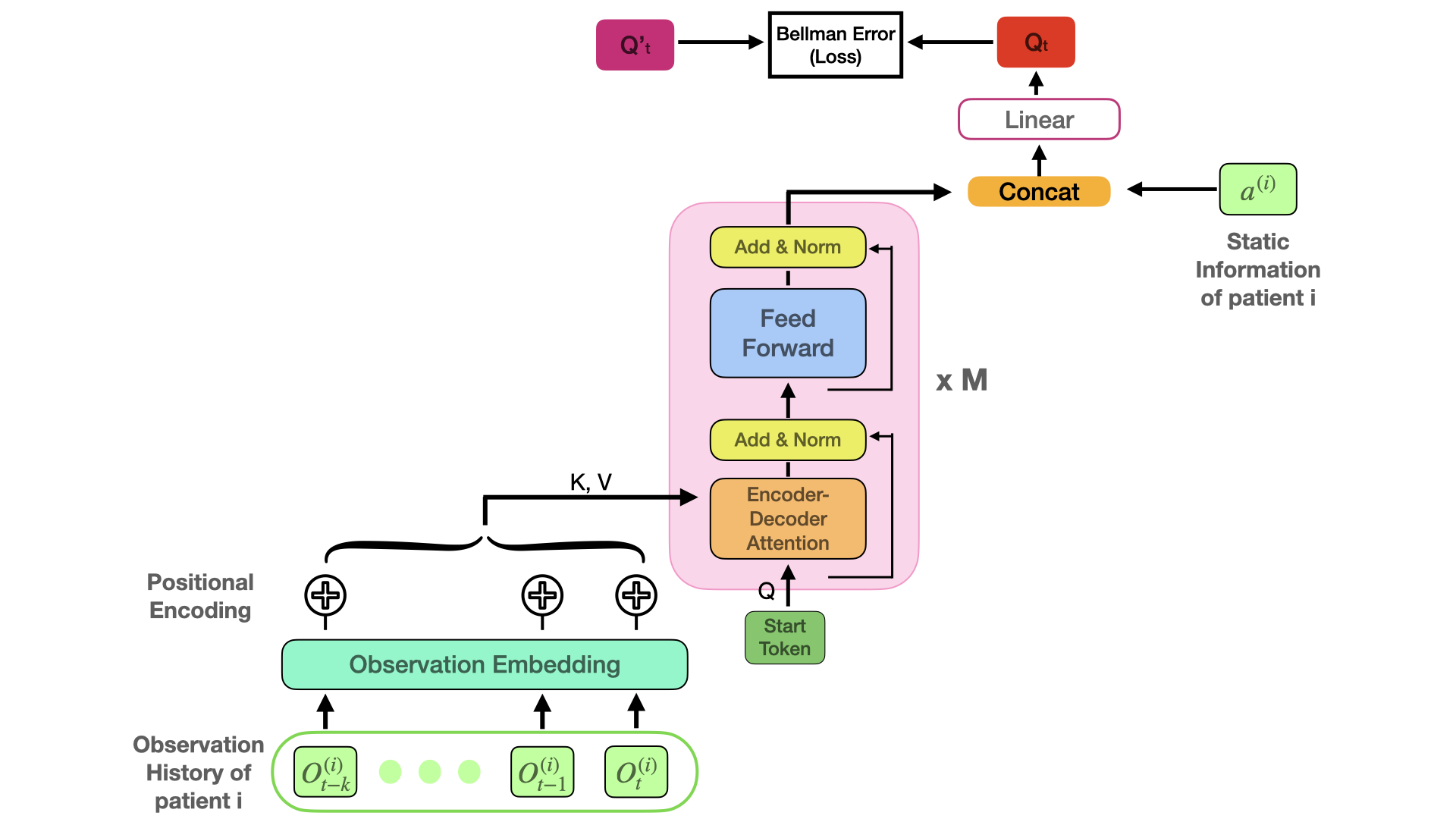}
\caption{Deep Attention Q-Network Architecture. Workflow: (1) input patient $i$ previous $k$ observations, $h_{t-k:t} = \{o_{t-k}^{(i)}, \ldots,o_{t-1}^{(i)}, o_t^{(i)}\}$; (2) linearly projects each observation into the model's dimension, and add learned positional encoding to each embedded observation based on its position in the observation history; (3) fed embedded and positional encoded observation through $M$ attention-like structure blocks, while attempting to decode to a fixed start token; (4) final output from $M$ attention-like structure blocks is concatenated with patient $i$'s static information vector $a^{(i)}$, and feed into a final linear layer to project into the action space dimension, $Q_t$, where each entry indicates the discounted cumulative reward given the current observation history and action of interest; (5) mean squared Bellman loss is computed \ref{eq:DuelDDQN_loss} between $Q_t$ and $Q'_t$ to learn all network parameters.}
\label{fig:DAQN_FlowChart}
\end{figure}

The transformer architecture \citep{vaswani2017attention}, originally introduced for sequence to sequence translation in natural language applications, utilizes attention mechanism \citep{bahdanau2014neural}, which is able to ``focus'' on different portions of the input when translating to outputs. With its strong interpretability and computational efficiency, the transformer architecture, originally formed as an encoder-decoder structure, is now broadly used in various applications, using either the encoder \citep{kenton2019bert}, the decoder \citep{yang2019xlnet}, or reconstructed architectures \citep{wu2021autoformer}. Transformer and its attention structures seem like a natural fit to represent the histories in POMDPs, as it encapsulates several inductive biases nicely. Therefore, we propose Deep Attention Q-Network (DAQN), by ``inserting'' the attention mechanisms to the traditional Deep Q-Networks to learn the approximate Q-function. The high-level overview of the proposed DAQN is shown in Figure \ref{fig:DAQN_FlowChart}, and the detailed workflow is presented in the caption.

Similar to the original Transformer's decoder structure \citep{vaswani2017attention}, each attention-like blocks in DAQN features two main submodules: encoder-decoder attention and position-wise feedforward network. First, in the encoder-decoder attention, the fixed start token (serving as a dummy variable) will be projected to queries $Q$, and positional encoded and embedded observation histories will be projected to keys $K$ and values $V$, through the learned weight matrices $W^Q, W^K, W^V$, respectively. Then, via ``Scaled Dot-Product Attention'' \citep{vaswani2017attention}, a softmax function is applied on the dot products between queries and keys, which are the attention weights and are used to obtain a weighted sum on the values. On the high-level, the attention weights can be interperated as the ``importance'' weight of each observation history relative to the the Q-values of state-action pairs. Then the output from the encoder-decoder attention will pass through a fully connected feed-forward network, which is applied to each position separately and identically. After each submodule, that submodule’s input and output are combined and followed by layer normalization \citep{ba2016layer}. 

We additionally incorporate Dueling Q-network architecture \citep{wang2016dueling} and Double-Deep Q-network architecture \citep{van2016deep} into our proposed architecture, and also use Prioritized Experience Replay \citep{schaul2015prioritized} to accelerate learning, similar to prior RL sepsis studies \citep{raghu2017continuous, raghu2017deep, peng2018improving}. More implementation details are presented in the Appendix \ref{sec:DAQN_more_detail}.

\section{Cohort}\label{sec:Cohort}
We obtain the data from the ``Medical Information Mart for Intensive Care database'' (MIMIC-III) version 1.4 \citep{johnson2016mimic}, a publicly available database consist of deidentified health-related data associated with patients stayed in critical care units of the Beth Israel Deaconess Medical Center between 2001 and 2012. The database includes information such as demographics, vital sign measurements made at the bedside, laboratory test results, procedures, medications, etc. In this study, we focus on two ICU patient cohorts: sepsis patient and acutely hypotensive patient.

\subsection{Sepsis Patients}
\subsubsection*{Data Extraction}
We extract all adult sepsis patients in MIMIC-III database that fulfill the following criteria: (1) patients who were older than 18 years old; (2) patients with the length of stay over 24 hours (to ensure sufficient data for analysis); (3) patients with the diagnosed of sepsis according to the Sepsis-3 criteria \citep{singer2016third}. Also, if a patient had multiple admissions with sepsis, only the first admission was analyzed. After excluding patients with relevant variables missing (see Table \ref{tab:Sepsis_Observ}), we have a total of 6,164 patients. For each patient, we extract the relevant physiological parameters including demographics, lab values, vital signs, and intake/output events (see Table \ref{tab:Sepsis_Observ}). Then, each patient's record is aggregated into windows of 4 hours, with the mean or sum being recorded when several data points were present in one window (same as prior studies \citep{komorowski2018artificial, raghu2017deep}). This yielded a feature vector for each patient at each timestep. For each patient $i\in\{1,\ldots,N\}$, at each timestep $t$, the current feature vector and the previous feature vectors (and previous actions) form the observation $o_t^{(i)}$ in the underlying POMDP.

\subsubsection*{Action and Reward}
We focus on the action space defined by two treatment interventions: intravenous fluid (IV fluid) and vasopressor, given their uncertainty in clinical literature \citep{napolitano2018sepsis} and the crucial impact on a patient's eventual outcome. We define a $5\times 5$ action space for the medical interventions covering the space of intravenous (IV) fluid and maximum dose of vasopressor in each 4-hour period (4 per-drug quartiles, and a special case of no drug), similar to prior studies' action space discretization \citep{raghu2017deep, komorowski2018artificial}. Thus, we are end up with total of 25 actions, each representing the intervention as a tuple of Input 4H and Max Vaso at each 4-hour period. 

To train an RL agent for sepsis management, we adopted a similar reward function as in \cite{raghu2017deep}, which uses the Sequential Organ Failure Assessment scores (SOFA) \citep{vincent1996sofa} and the lactate level of the patients. On the high level, higher SOFA scores indicate greater organ dysfunction and is predictive of ICU mortality, while lactate levels measure cell-hypoxia which is higher in septic patients. The rewards function penalizes high SOFA scores and lactate levels at time $t$, as well as positive changes in these scores. Conversely, positive rewards are given for decreased SOFA scores and lactate levels, indicating improved patient states. Further details on the reward function are in Appendix \ref{sec:Feature_Rewards}.

\subsection{Acutely Hypotensive Patients}
\subsubsection*{Data Extraction}
Following prior study \citep{gottesman2020interpretable}, we extract the acutely hypotensive patients in MIMIC-III database that fulfill the following criteria: (1) patients who were older than 18 years old; (2) patients with the length of stay over 24-hours (and select the initial ICU admission only); (3) patients with seven or more mean arterial pressure (MAP) values of 65 mmHg or less, which indicated probable acute hypotension. For each patient, we extract the relevant physiological parameters (Table \ref{tab:AcuteHypotension_Observ}), and limit to only using information captured during the initial 48 hours after admission. We have a final cohort consisting of 3910 distinct ICU admissions.

\subsubsection*{Action and Reward}
Following prior studies \citep{gottesman2020interpretable}, we choose to focus on the action space defined by two treatment interventions: fluid boluses and vasopressor, defining a $4\times 4$ action space (3 per-drug quartiles, and a special case of no drug). 

We adopted a similar reward function as \cite{gottesman2020interpretable} for training an RL agent for the management of acute hypotension. The reward at time step $t$ of a patient, is dependent on the Mean Arterial Pressure (MAP) and urine output at time $t$. The detailed formula is provided in Appendix \ref{sec:Feature_Rewards}.

\section{Results} 
In this section, we show how the proposed DAQN dynamically suggest optimal personalized healthcare treatments, in intensive care units (ICU). In this study, we focus on off-policy learning, which means that our RL agent aims to learn an optimal policy (i.e. optimal medication dosages) through data that are already generated by following the clinician policy (see section \ref{sec:Cohort}). For each cohort, we learn the optimal DAQN policy, while conduct evaluation comparisons against other benchmark policies.

Proper quantitative evaluation of learned policy is crucial before deployment, especially in healthcare. Off-policy evaluation (OPE) in the reinforcement learning context is typically used as the performance metric for comparison \citep{thomas2016data, jiang2016doubly}. Here, we employ weighted doubly-robust method (WDR) to quantify the performance of RL policies \citep{thomas2016data}. We include the proposed DAQN policy, DRQN (Deep Recurrent Q-Network) policy \citep{hausknecht2015deep}, DQN (Deep Q-Network) policy \citep{raghu2017deep}, clinician policy, and random policy for comparison. The DRQN and DQN policy use vanilla LSTM network (long-short-term-memory) \citep{hochreiter1997long} and feedforward neural networks \citep{mnih2015human} for learning the approximate Q-function (see Equation \ref{eq:Q_update}), respectively. The clinician policy comprises actions from historical data which clinicians take. For random policy, actions is uniformly sampled from the 0 to safety upper bound range. More details on the benchmark policies are presented in the Appendix \ref{sec:BenchmarkMethod}. To account for randomness, we performed 50 experiments with a unique train/test set split in each experiment.

\subsection*{Sepsis Patients}
\begin{table}[!ht]
\sisetup{detect-weight,mode=text}
\renewrobustcmd{\bfseries}{\fontseries{b}\selectfont}
\renewrobustcmd{\boldmath}{}
\newrobustcmd{\B}{\bfseries}
\centering
\begin{tabular}{lrr}
\hline
Evaluating Policies & Sepsis & Acute Hypotension \\ \hline
DAQN & \textbf{0.34893$\pm$0.0632} & \textbf{-0.00562$\pm$0.0033} \\
DRQN & 0.23683$\pm$0.0708  &  -0.01014$\pm$0.0052\\
DQN &  0.16710$\pm$0.0646  & -0.01422$\pm$0.0051\\
Clinician & 0.06772$\pm$0.0149  & -0.02363$\pm$0.0014\\
Random &  -0.03517$\pm$0.0617 & -0.03857$\pm$0.0073 \\\hline
\end{tabular}
\caption{Mean and standard deviation of off-policy evaluation estimates via WDR estimator \citep{thomas2016data}, across evaluating policies. The best performing policy is boldfaced.} \label{tab:OPE}
\end{table}

The results are presented in Table \ref{tab:OPE} and the box plot in Figure \ref{fig:OPE_Sepsis}. The quantitative results demonstrate that our proposed DAQN policy is able outperform benchmark DRQN policy in both mean and standard deviation, which also incorporates historical patients' observations for state representation, while outperforming traditional DQN policy, clinician policy, and random policy. The value of the clinician's policy is estimated with high confidence, as expected. DAQN and DQN-based policy all bound on the estimated value relatively tight and larger than the value of the clinician's policy (1st to 3rd quartile box small and above the mean for the clinician’s policy), in contrast to the random policy's large box extending to well below the clinician's policies. The random policy's values are distributed evenly around zero, which is expected as the reward distributions is also approximately distributed around zero. DRQN policy is also able to outperform clinician's policy and baseline random policy, and exhibit larger mean than DQN policy, but exhibit the largest variance. 

\begin{table}[!ht]
\sisetup{detect-weight,mode=text}
\renewrobustcmd{\bfseries}{\fontseries{b}\selectfont}
\renewrobustcmd{\boldmath}{}
\newrobustcmd{\B}{\bfseries}
\addtolength{\tabcolsep}{-4.1pt}
\footnotesize
\centering
\begin{tabular}{l|r|r|r}
\hline
Attention Layer & SOFA & Delta SOFA &Lactate\\\hline
Layer 1 & 0.626 & 0.368 & 0.209\\
Layer 2 & 0.579 & 0.723 & 0.340\\
Layer 3 & 0.390 & 0.120 & 0.575\\
Layer 4 & 0.340 & 0.383 & 0.251\\\hline
\end{tabular}
\caption{Correlation coefficient between averaged attention weights (across each head) in each layer and SOFA score, Delta SOFA score, and lactate level.} \label{tab:Sepsis_Attn_corrcoef}
\end{table}

In addition, we examine the interpretability in our proposed DAQN policy, by focusing on the attention weights over the input historical patient observations. We observe that the attention weights, produed in the ``Encoder-Decoder Attention'' block (in Figure \ref{fig:DAQN_FlowChart}), is positively associated with patient's SOFA score, change in SOFA score, and lactate level, which are all important indicator in sepsis patients, with high association with mortality and morbidity \citep{lambden2019sofa}. Numerous studies \citep{liu2019prognostic, vincent1996sofa,lambden2019sofa, ferreira2001serial} show that the SOFA score is highly sensitive and predictive in the diagnosis of sepsis. For instance, a initial and highest scores of more than 11 or mean scores of more than 5 corresponded to mortality of more than 80\%, while change in SOFA score (delta SOFA score) also significantly associated with mortality and patient discharge \citep{ferreira2001serial}. Indeed, with the designed reward function that penalizes high SOFA scores, positive changes in SOFA scores, and high lactate levels, the attention weights learn to ``focus'' on the prior observations and actions that exhibit high SOFA score, change in SOFA score, or lactate level, respectively. Table \ref{tab:Sepsis_Attn_corrcoef} shows the correlation coefficient between the average attention weights (across attention heads) in each layer and SOFA score, change in SOFA score, lactate level, respectively. Then, we select three example patients and visualize the average attention weights in each layer with SOFA score, delta SOFA score, and lactate level (the elements that has the highest correlation coefficient with each layer's average attention weight, see Table \ref{tab:Sepsis_Attn_corrcoef}), in Figure \ref{fig:ExamplePatient1_AttnWeights}, \ref{fig:ExamplePatient2_AttnWeights}, \ref{fig:ExamplePatient3_AttnWeights}. For example, the attention weights of example patient 1 in \ref{fig:ExamplePatient1_AttnWeights} exhibit strong trend matching behavior with SOFA score, Delta SOFA score, and lactate level, and learn to ``focus'' (high attention weights) on past observations that indicates worse patient's health status. This further confirms DAQN's ability to focus on more ``important'' and severe historical observations when learning the optimal policy. More details on presented in Appendix \ref{sec:more_results}.

\begin{figure}[htbp]
\floatconts {fig:ExamplePatient1_AttnWeights}
{\caption{Example sepsis patient 1 average attention weights in each layer against SOFA score, delta SOFA score, and lactate level, respectively. 
}}
{%
\subfigure[SOFA score and attention weights (layer 1)]{%
  \label{fig:Patient1_Layer1}
  \includegraphics[width=0.4\textwidth]{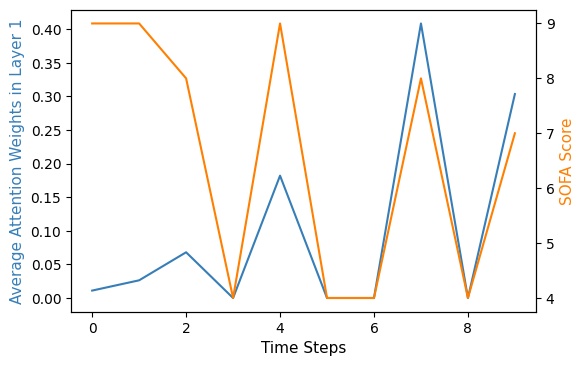}
}\qquad 
\subfigure[Delta SOFA score and attention weights (layer 2)]{%
  \label{fig:Patient1_Layer2}
  \includegraphics[width=0.4\textwidth]{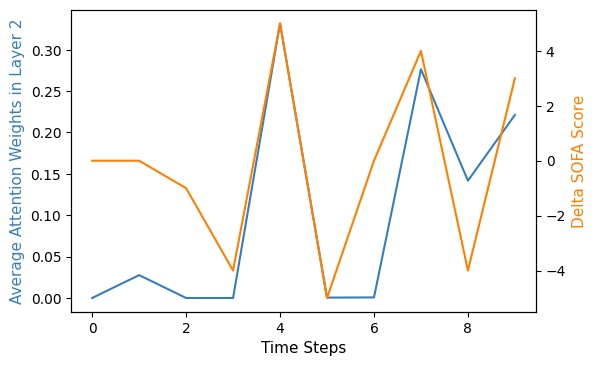}
}\qquad 
\subfigure[Lactate level (normalized) and attention weights (layer 3)]{%
  \label{fig:Patient1_Layer3}
  \includegraphics[width=0.4\textwidth]{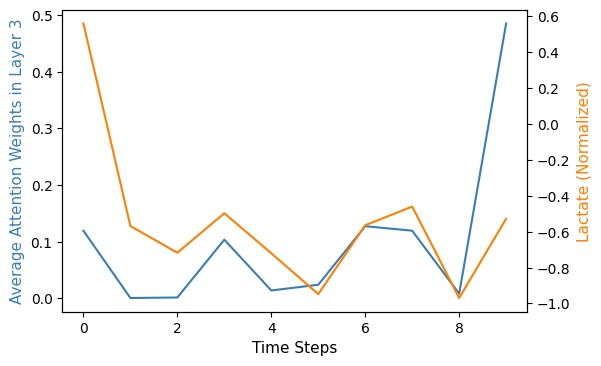}
}\qquad 
\subfigure[Delta SOFA score and attention weights (layer 4)]{%
  \label{fig:Patient1_Layer4}
  \includegraphics[width=0.4\textwidth]{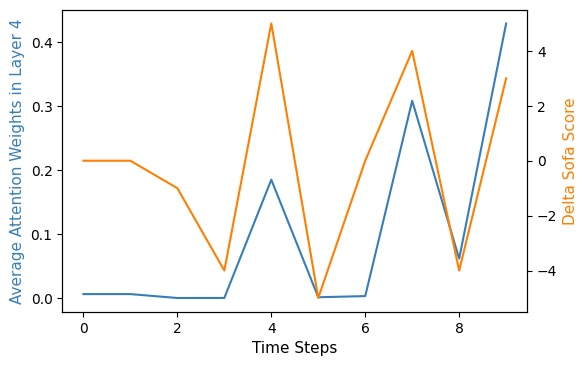}
}
}
\end{figure}

\subsection*{Acutely Hypotensive Patients}
Managing hypotensive patients in the ICU is a challenging task that lacks standardization due to the high heterogeneity of patients, which often leads to high morbidity and mortality rates \citep{jones2006emergency}. In light of the limited evidence to guide treatment guidelines, RL offers a promising approach to improve strategies for managing these patients \citep{de2018unexplained}. To evaluate the efficacy of RL policies for acutely hypotensive patients, we conducted experiments on the MIMIC-III dataset, and the results are presented in Table \ref{tab:OPE} and the box plot in Figure \ref{fig:OPE_AcuteHypotension}. We used the same hyperparameters for DAQN and the benchmark policies as in the previous section. Our results show that, similar to the sepsis patient cohort, the DAQN policy outperforms the DRQN and DQN policies, with the 3rd quartile box located above the mean of the DRQN and DQN policies. This performance improvement is attributed to the Transformer architecture and attention mechanism employed by the DAQN, which enables it to focus on and efficiently memorize past patient observations and actions as current patient health status representation more robustly than the DRQN, which can suffer from vanishing/exploding gradient problems due to RNNs. However, the DRQN policy shows a higher mean than the DQN policy, which underscores the importance of modeling the ICU setting as a POMDP problem, treating the current patient vitals and static information as observations, and not as the ``true'' patient underlying health status. The estimated values of the DAQN, DRQN, and DQN policies are relatively tight, and larger than those of the clinician policy and randomized policy. The expected reward of the clinician policy is estimated with high confidence, while the randomized policy's reward distribution is similar to that of the entire cohort. Overall, our results demonstrate the potential of RL policies, especially the DAQN, to improve the management of acutely hypotensive patients in the ICU.

\section{Discussion}
Various pioneering studies have explored applying reinforcement learning algorithms to the search for optimal clinical treatment, such as for sepsis patients \citep{raghu2017continuous, peng2018improving, raghu2017deep, huang2022reinforcement} and for acutely hypotentive patients \citep{gottesman2020interpretable}, demonstrating the potential of using RL to improve ICU patient outcomes. However, these studies are limited to a coarse-grained state space that only depends on the current patient's observations for state space definitions. Although this representation of patient's state space is intuitive and straightforward for determining current actions, patients' prior observations and prior actions are also important factors of treatment intervention decisions. Thus, simply modeling the patient-clinician interactions as MDPs could prone to mis-specification of the ``true'' underlying patient's states, and potentially take sub-optimal actions, leading to sub-optimal results (rewards). In order to make personalized treatment design more clinically meaningful, we proposed to ``enrich'' the state space, by modeling the ICU setting as a POMDP and letting the RL agent efficiently memorize patients' current and prior observations and actions. By incorporating attention mechanism, our proposed RL algorithm is able to outperform baseline benchmark policies, and provide interpretability, similar to clinician's diagnosis process. 

By extending state space with prior observations, we learn the optimal policy using our proposed RL algorithm, which can provide more meaningful and higher-resolution decision support to patients. For quantitative policy evaluation, we compare our optimal policy with other policies learned by alternative benchmark RL algorithms, as well as baseline policies, through off-policy evaluation with WDR estimator \citep{thomas2016data}. The results from off-policy evaluations shows that the proposed the RL policy is able to perform competitively against alternative benchmark policy that uses simpler memorization techniques (LSTM), and outperform other RL policies that do not incorporate prior observations (and actions). The proposed RL policy can also provide better expected reward compared to clinician policy and random policy baseline. Moreover, similar to clinician decision process and thinking, the proposed RL policy will focus on the prior observations (and actions) that indicate worse patient health in combination of current observations, when making  treatment dosage decisions (see Figure \ref{fig:ExamplePatient1_AttnWeights}, \ref{fig:ExamplePatient2_AttnWeights}, \ref{fig:ExamplePatient3_AttnWeights}). With clinician guided reward function and attention mechanism, our proposed RL policy is able to shift focus among previous patient's observations for subsequent treatment decisions. This improvement also makes reinforcement learning-based treatment search closer to real-world deployment.

Potential avenues for future work include a more thorough discussion with clinicians to potentially make the observation and action histories even more representative, and architectural improvements that could provide more detailed interpretation for patient-intervention relation.

\paragraph{Limitations} 
As mentioned by prior studies \cite{raghu2017deep, gottesman2018evaluating}, evaluating RL policies with off-policy evaluation is challenging, as all the available data are offline sampled (i.e. following clinician's policy). Since the evaluation policies are deterministic, the the off-policy evaluation that uses importance weight will only be non-zero if the evaluation policy recommends the same treatment as the clinician's policy. The will results in high-variance estimates of the quality of the evaluation policy. Future examination from both policy learning and policy evaluation aspects shall be considered. 

Another limitation of this study is that the dosages are discretized into per-drug quartiles, with each action representing dosages in a particular range. However, each quartile includes a wide range of dosages , which can be complicated in practice for clinicians to make decisions on the exact dosages of IV fluids and vasopressor to use. \cite{huang2022reinforcement} proposed a RL-based solution via Deep Deterministic Policy Gradient \citep{lillicrap2015continuous}, which is a model-free off-policy algorithm for learning continuous actions. Therefore, it is also important to further investigate policy-gradient algorithms that efficiently memorizes patient's prior observations and actions continuous actions.


\bibliography{references}

\clearpage
\appendix
\section{}
\subsection{Implementation Details}\label{sec:Implementation_detail}

\subsubsection{Propsoed DAQN modifications} \label{sec:DAQN_more_detail}
\begin{figure}[htbp]
\centering
\includegraphics[width=0.8\textwidth]{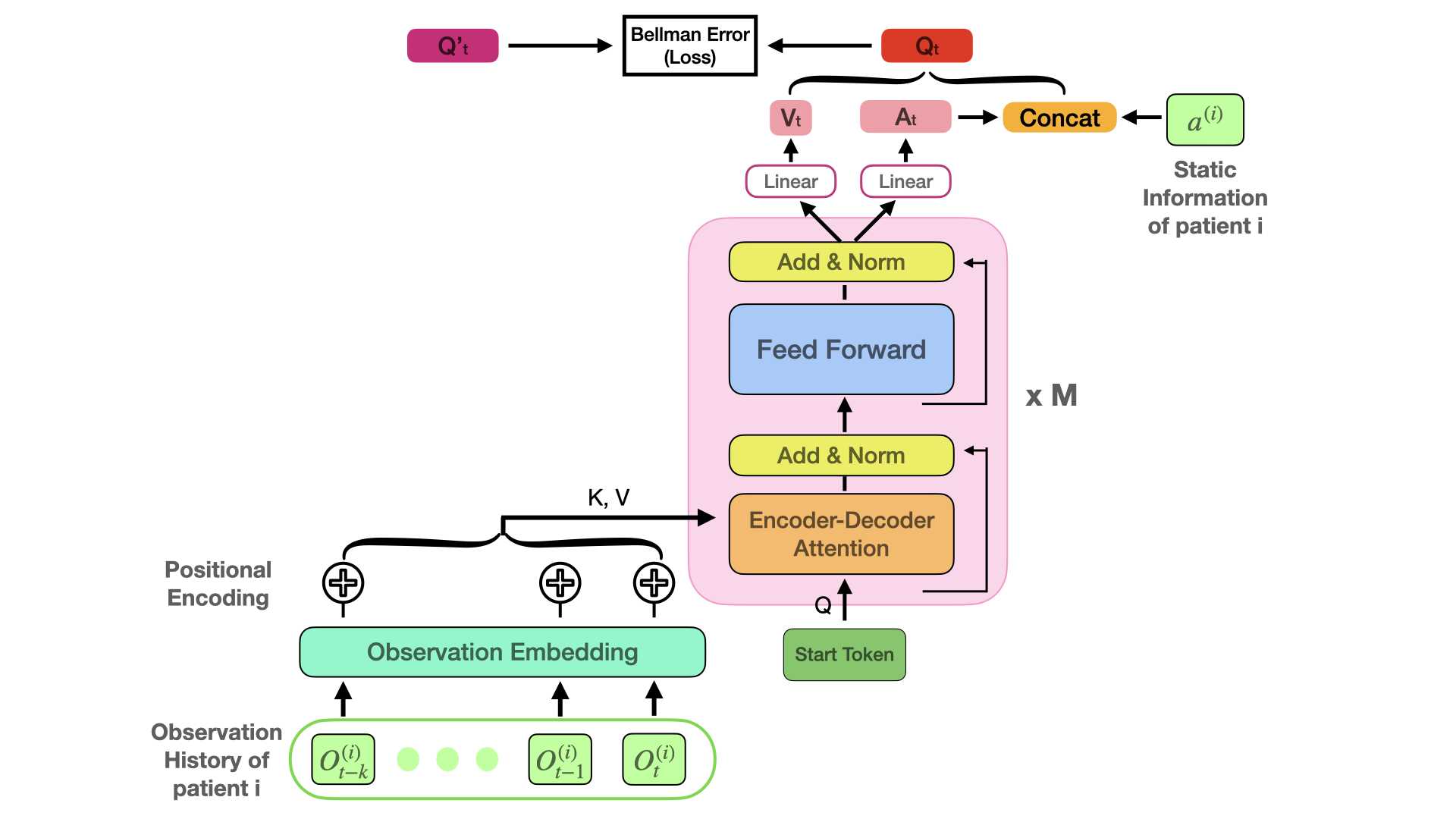}
\caption{Proposed Dueling-Double-Deep-Attention-Q-Network Architecture (Dueling-DDAQN)}.
\label{fig:DuelDDAQN_FlowChart}
\end{figure}

Similar to prior RL studies on sepsis \citep{raghu2017continuous, raghu2017deep, peng2018improving}, we incorporate Dueling Q-network architecture \citep{wang2016dueling}, and 
Double-Deep Q-network architecture \citep{van2016deep} into our proposed DAQN, to combat a few shortcomings of traditional Q-Networks. Dueling Q-network architecture \citep{wang2016dueling} proposed to produce separate state-value ($V_t$) and action advantages ($A_t$) streams, through two different liner layers, instead of a single linear layer (see Figure \ref{fig:DAQN_FlowChart}). This will separate the effect on the Q-values of a patient being in a good underlying state from a good action being taken, improving the learning \citep{wang2016dueling}). Lastly, the state-value and action advantages will combine into a set of Q-values, in action space dimension, $Q_t$. Double-Deep Q-network architecture \citep{van2016deep} helps correct overestimation of Q-values by using a second target network to compute the Q-values, i.e. using two different networks for maximum calculation and Q-value estimation in the mean squared Bellman error (see Equation \ref{eq:DQN_loss}). Thus, we denote our proposed method as Dueling-Double-Deep-Attention-Q-Network (Dueling-DDAQN), and shown in the flow chart in Figure \ref{fig:DuelDDAQN_FlowChart}.

Table \ref{tab:Hyperparameter} hyperparameter combinations used for the Dueling-DDAQN architecture thorough out the experiments .
\begin{table}[htbp]
\sisetup{detect-weight,mode=text}
\renewrobustcmd{\bfseries}{\fontseries{b}\selectfont}
\renewrobustcmd{\boldmath}{}
\newrobustcmd{\B}{\bfseries}
\addtolength{\tabcolsep}{-4.1pt}
\footnotesize
\centering
\begin{tabular}{ll}
  \hline
Hyperparameter & Value\\ \hline
No. attention blocks & 4\\
No. attention head & 2\\
Embedding Dimension & 128\\
Feed Forward Dimension & 256\\\hline
\end{tabular}
\caption{Hyperparameter of Dueling-DDAQN} \label{tab:Hyperparameter}
\end{table}

\subsubsection{Benchmark methods details} \label{sec:BenchmarkMethod}
The Deep Recurrent Q-Network (DRQN) structure used in evaluation comparisons uses a vanilla LSTM structure, with hidden unit and cell unit being in 128 dimension \citep{hochreiter1997long}. The Deep Q-Network (DQN) structure used in evaluation comparison uses two feedforward neural network, with dimensionality of 128. Both DRQN and DQN also incorporate the Dueling Q-network architecture  \citep{wang2016dueling}, and 
Double-Deep Q-network architecture \citep{van2016deep} for fair comparisons, i.e. denoting as Dueling-DDRQN and Dueling DDQN. Also, for both DAQN and DRQN, we use a look back window of 9, i.e. incorporate up to $k=9$ historical patient observation as the current state representation (see \ref{fig:DuelDDAQN_FlowChart}).

The final mean-squared Bellman error loss function used for learning DTQN, DRQN and DQN policies is as follows:
\begin{equation}
L(\theta) = \mathbb{E}_{s,a,r,s'}\Bigg[\left(r+\gamma Q(s', \argmax_{a'\in\mathcal{A}}Q(s',a';\theta);\theta') - Q(s,a;\theta)\right)^2\Bigg]
\label{eq:DuelDDQN_loss}
\end{equation}
where $\theta$ are the weights used to parameterize the main network, and $\theta'$ are the weights used to parameterize the target network. We use a train/test split of 80/20 and ensure that a proportionate number of patient outcomes are present in both sets.  During training, we sample transitions of the form $\{s, a, r, s'\}$ from our training set, using the Prioritized Experience replay scheme \citep{schaul2015prioritized}, perform feed-forward passes on the main and target networks to evaluate the output and loss, and update the weights in the main network via backpropagation. Training was conducted for 10000 batches, with batch size 128. The reward discount factor for DAQN, DRQN and DQN policy are set to be $\gamma=0.99$ (see Equation \ref{eq:DQN_loss}).

\subsection{More details on problem setup and data pre-processing Functions}\label{sec:Feature_Rewards}
\subsubsection{Selected Features}
\subsubsection*{Sepsis}
For all sepsis patients, we include 42 dynamic variables and 5 static variables as observations, to define a patient's state in each encounter (4-hour period). Among the variables, IV fluids and vasopressors define the action space and the rest define the state space, in the RL setup. The dynamic variables are shown in Table \ref{tab:Sepsis_Observ}. The static variables are: gender, mechanical ventilation, readmission, age, and weight.

\begin{table}[htbp]
\sisetup{detect-weight,mode=text}
\renewrobustcmd{\bfseries}{\fontseries{b}\selectfont}
\renewrobustcmd{\boldmath}{}
\newrobustcmd{\B}{\bfseries}
\addtolength{\tabcolsep}{-4.1pt}
\scriptsize
\centering
\begin{tabular}{lll}
  \hline
Variable Name & Data Type & Unit\\\hline
Albumin&numeric&g/dL\\
Arterial Base Excess (BE)&numeric&meq/L\\
Bicarbonate (HCO3)&numeric&meq/L\\
Creatinine&numeric&mg/dL\\
Diastolic BP &numeric &mmHg\\
FiO2&numeric&fraction\\
Glucose&numeric&mg/dL\\
Heart Rate (HR) & numeric & bpm\\
Intravenous Fluids of Each 4-Hour Period (Input 4H)&numeric&mL\\
Lactate&numeric&mmol/L\\
Mean BP	& numeric &mmHg\\
Systolic BP	& numeric &mmHg\\
Respiratory Rate (RR) &numeric &bpm\\
Potassium (K+) &numeric &meq/L\\
Sodium (Na+) &numeric &meq/L\\
Chloride (Cl-) &numeric &meq/L\\
Calcium (Ca++) &numeric	&mg/dL\\
Ionised Ca++&numeric &mg/dL\\
Carbon Dioxide (CO2)&numeric&meq/L\\
Hemoglobin (Hb)&numeric&g/dL\\
Potential of Hydrogen (pH)&numeric & -\\
Blood Urea Nitrogen (BUN)&numeric&mg/dL\\
Magnesium (Mg++)&numeric&mg/dL\\
Serum Glutamic Oxaloacetic Transaminase (SGOT) &numeric&u/L\\
Serum Glutamic Pyruvic Transaminase (SGPT)&numeric&u/L\\
Total Bilirubin (Total Bili)&numeric&mg/dL\\
White Blood Cell Count (WBC)&numeric&E9/L\\
PaO2&numeric&mmHg\\
Partial Pressure of CO2 (PaCO2)&numeric&mmHg\\
Platelets Count (Platelets)&numeric&E9/L\\
Total Volume of Intravenous Fluids (Input Total)&numeric&mL\\
Total Volume of Urine Output (Output Total)&numeric&mL\\
Maximum Dose of Vasopressors in 4H (Max Vaso)&numeric&mcg/kg/min\\
Urine Output in 4H (Output 4H)&numeric&mL\\
Glasgow Coma Scale	& categorical&\\
Pulse Oximetry Saturation (SpO2)& categorical &-\\
Temperature (Temp)& categorical & Celcius\\
Partial Thromboplastin Time (PTT) & categorical &s\\ 
Prothrombin Time (PT) &categorical &s\\
International Normalised Ratio (INR) &categorical&-\\\hline
\end{tabular}
\caption{Sepsis patient's observations.} \label{tab:Sepsis_Observ}
\end{table}

\subsubsection*{Acute Hypotension}
We include 22 dynamic variables and 5 static variables for acutely hypotensive patients, see Table \ref{tab:AcuteHypotension_Observ}. The static variables are: gender, mechanical ventilation, readmission, age, and weight.

\begin{table}[htbp]
\sisetup{detect-weight,mode=text}
\renewrobustcmd{\bfseries}{\fontseries{b}\selectfont}
\renewrobustcmd{\boldmath}{}
\newrobustcmd{\B}{\bfseries}
\addtolength{\tabcolsep}{-4.1pt}
\scriptsize
\centering
\begin{tabular}{lll}
  \hline
Variable Name & Data Type & Unit\\\hline
Alanine Aminotransferase (ALT)  & numeric  & IU/L\\
Aspartate Aminotransferase (AST)  & numeric  & IU/L\\
Diastolic Blood Pressure (Diastolic BP)  & numeric  & mmHg\\
Mean Arterial Pressure (MAP) & numeric  & mmHg\\
Systolic Blood Pressure (Systolic BP)  & numeric  & mmHg\\
Urine  & numeric  & mL\\
Partial Pressure of Oxygen (PaO2)  & numeric  & mmHg\\
Lactate  & numeric &  mmol/L\\
Serum Creatinine  & numeric  & mg/dL\\
Fluid Boluses  & categorical  & mL\\
Vasopressors  & categorical  & mcg/kg/min\\
Fraction of Inspired Oxygen (FiO2)  & categorical  & fraction\\
Fluid Boluses  & categorical  & mL\\
Glasgow Coma Scale Score (GCS)  & binary  & -\\
Urine Data Measured (Urine (M))  & binary  & -\\
ALT or AST Data Measured (ALT/AST (M))  & binary  & -\\
FiO2 (M)  & binary  & -\\
GCS (M)  & binary  & -\\
PaO2 (M)  & binary  & -\\
Lactic Acid (M)	&binary	-\\
Serum Creatinine (M)&binary	&-\\\hline
\end{tabular}
\caption{Acutely Hypotensive patient's observations.} \label{tab:AcuteHypotension_Observ}
\end{table}

\subsubsection{Reward Function}
\subsubsection*{The Sepsis Reward Function}
We adopted the reward function from \cite{raghu2017deep} for training an RL agent for the
management of sepsis. The reward function is computed in three parts such that:
\begin{equation}
\begin{aligned}
&\mathrm{reward}_t = \mathrm{reward}^{(1)}_t + \mathrm{reward}^{(2)}_t+ \mathrm{reward}^{(3)}_t &\text{where}\\
&\mathrm{reward}^{(1)}_t  = -0.025 &\text{if $\mathrm{SOFA}_t=\mathrm{SOFA}_{t-1} \text{ and } \mathrm{SOFA}_t>0$}\\
&\mathrm{reward}^{(2)}_t  = -0.125(\mathrm{SOFA}_t-\mathrm{SOFA}_{t-1})\\
&\mathrm{reward}^{(3)}_t = -2\mathrm{tanh}(\mathrm{Lactate}_t-\mathrm{Lactate}_{t-1})
\end{aligned}
\end{equation}
Note that the SOFA score can be easily derived from the patient variables: PaO2, FiO2, Platelets, Total Bili, Mean BP, Max Vaso, GCS, Creatinine, and Output 4H (see Tables \ref{tab:Sepsis_Observ} for details of the above variables).

\subsubsection*{The Acute Hypotension Reward Function}
We adopted the reward function from \cite{gottesman2020interpretable} for training an RL agent for
the management of acute hypotension. The reward at time step $t$ is dependent on the Mean Arterial Pressure $\mathrm{MAP}_t$
and is given as:

\begin{equation}
\begin{aligned}
&\mathrm{reward}_t = \begin{cases} 0, &\text{$\mathrm{MAP}_t > 65$}\\ 
\frac{-0.05 (65-\mathrm{MAP}_t )}{5} ,&\text{$60<\mathrm{MAP}_t \leq 65$}\\
\frac{-0.1(60-\mathrm{MAP}_t)}{5} - 0.05, &\text{$55<\mathrm{MAP}_t \leq 60$}\\
\frac{-0.85(55-\mathrm{MAP}_t)}{15} - 0.15, &\text{$\mathrm{MAP}_t \leq 55$}
\end{cases}
\end{aligned}
\end{equation}
but the reward value is also dictated, and overwrote, by the Urine output $\mathrm{urine}_t$ of a patient when
\begin{align*}
\mathrm{reward}_t = 0, \;\;\text{if $\mathrm{urine}_t > 30 \text{ and } \mathrm{MAP}_t{5} > 55$}
\end{align*}

\subsection{More Results}\label{sec:more_results}
\subsubsection{Evaluation Results}
\begin{figure}[htbp]
\floatconts {fig:OPE}
{\caption{Policy value estimates via WDR estimator for sepsis and acutely hypotensive patients, for the evaluating policies. All boxes represent results over 50 random partitionings of the data into training (80\%) and testing (20\%) set.}}
{%
\subfigure[Sepsis Patients]{%
  \label{fig:OPE_Sepsis}
  \includegraphics[width=0.4\textwidth]{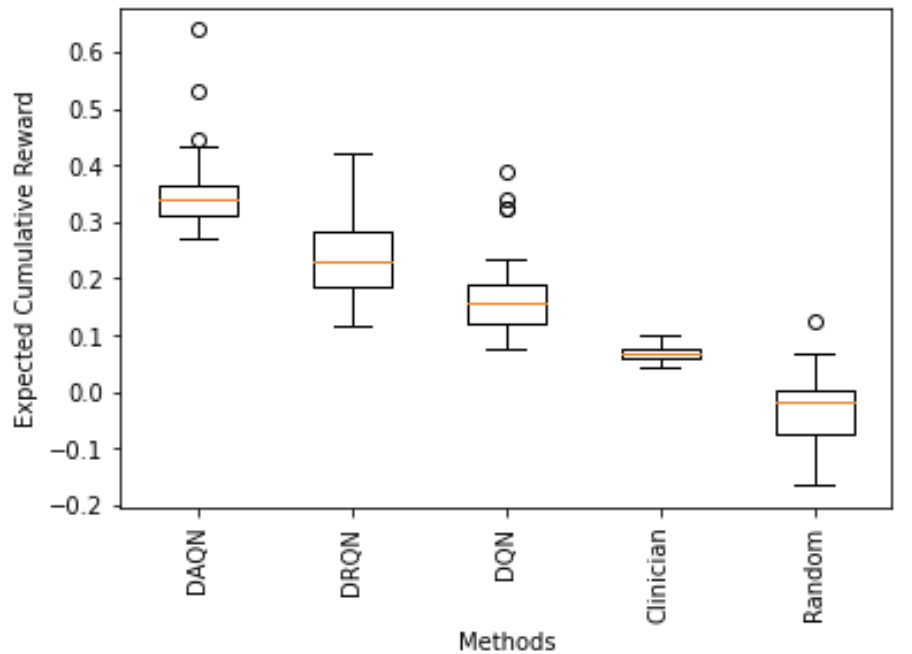}
}\qquad 
\subfigure[Acutely Hypotensive Patients]{%
  \label{fig:OPE_AcuteHypotension}
  \includegraphics[width=0.4\textwidth]{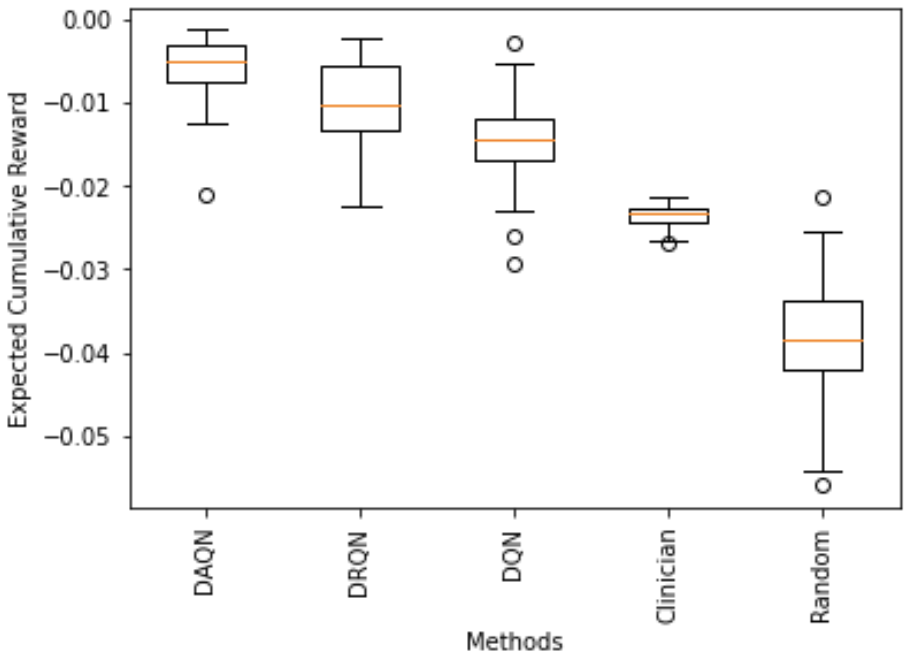}
}
}
\end{figure}

In this section, we present more evaluation results. Figure \ref{fig:OPE_Sepsis} and \ref{fig:OPE_AcuteHypotension} show the evaluating policies' value estimations via WDR estimator \citep{thomas2016data}, for sepsis patients and  acutely hypotentive patients respectively.

\subsubsection{Interpretability}
In addition to off-policy evaluations, our proposed DAQN policy is able to ``foucs'' on patients' different historical observations, when estimating the Q-values. To further investigate and visualize this, we first compute a the correlation coefficient of averaged attention weights in each layer (across each attention heads), with the SOFA score, delta SOFA score, and lactate, respectively. For example, for layer 1, we obtain the averaged attention weights across two attention heads (see hyperparameter choices in Table \ref{tab:Hyperparameter}), and compute its correlation coefficient with all the historical observations' SOFA score, delta SOFA score, and lactate. Additionally, we select three example patients and visualize the average attention weights in each layer and SOFA score, delta SOFA score, and lactate level, respectively (Figure \ref{fig:ExamplePatient1_AttnWeights}, \ref{fig:ExamplePatient2_AttnWeights}, \ref{fig:ExamplePatient3_AttnWeights}). Here, all three patients' observation sequence are select towards the end point of their records, i.e. the last time step in the visualization is also the end of their ICU records. 

\begin{figure}[htbp]
\floatconts {fig:ExamplePatient2_AttnWeights}
{\caption{Example sepsis patient 2's average attention weights in each layer against SOFA score, delta SOFA score, and lactate level, respectively. 
}}
{%
\subfigure[SOFA score and attention weights (layer 1)]{%
  \label{fig:Patient2_Layer1}
  \includegraphics[width=0.4\textwidth]{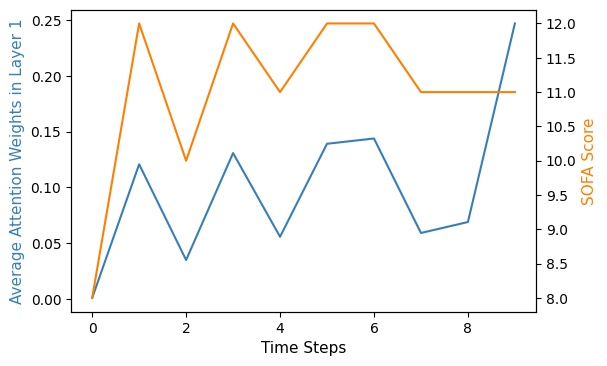}
}\qquad 
\subfigure[Delta SOFA score and attention weights (layer 2)]{%
  \label{fig:Patient2_Layer2}
  \includegraphics[width=0.4\textwidth]{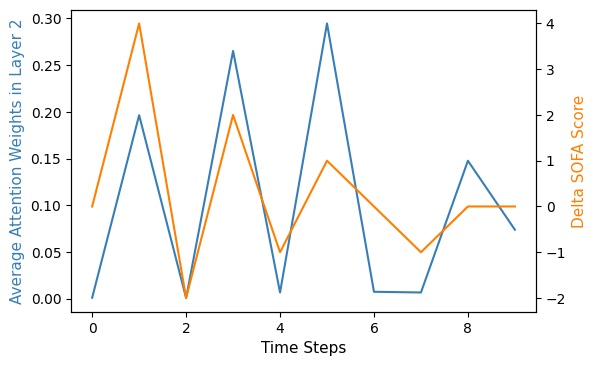}
}\qquad 
\subfigure[Lactate level (normalized) and attention weights (layer 3)]{%
  \label{fig:Patient2_Layer3}
  \includegraphics[width=0.4\textwidth]{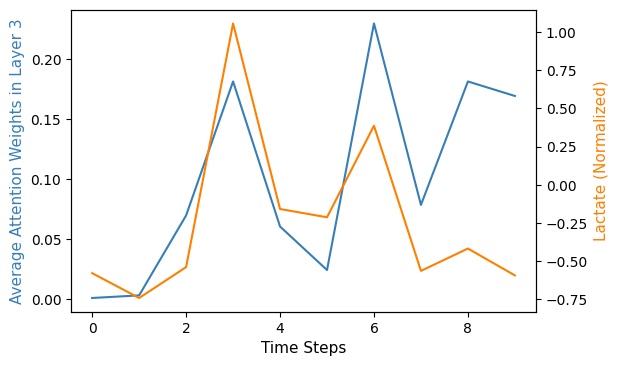}
}\qquad 
\subfigure[Delta SOFA score and attention weights (layer 4)]{%
  \label{fig:Patient2_Layer4}
  \includegraphics[width=0.4\textwidth]{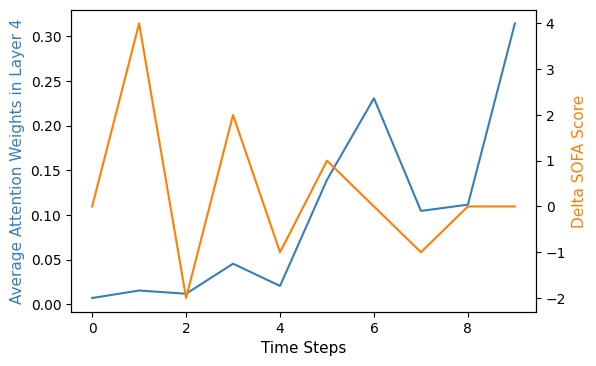}
}
}
\end{figure}

\begin{figure}[htbp]
\floatconts {fig:ExamplePatient3_AttnWeights}
{\caption{Example sepsis patient 3's average attention weights in each layer against SOFA score, delta SOFA score, and lactate level, respectively. 
}}
{%
\subfigure[SOFA score and attention weights (layer 1)]{%
  \label{fig:Patient3_Layer1}
  \includegraphics[width=0.4\textwidth]{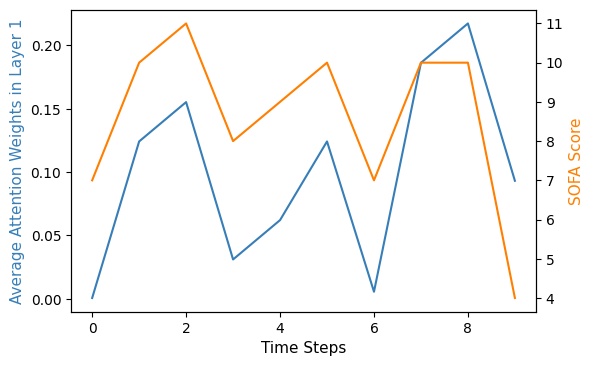}
}\qquad 
\subfigure[Delta SOFA score and attention weights (layer 2)]{%
  \label{fig:Patient3_Layer2}
  \includegraphics[width=0.4\textwidth]{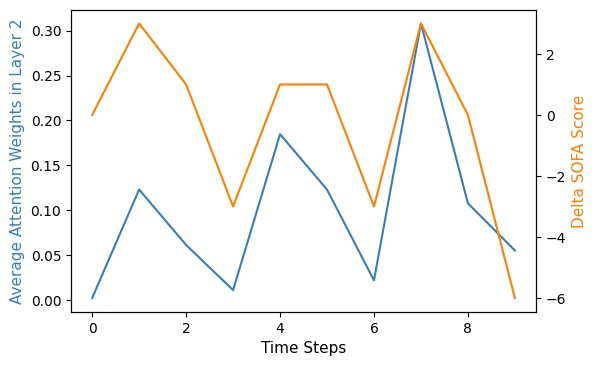}
}\qquad 
\subfigure[Lactate level (normalized) and attention weights (layer 3)]{%
  \label{fig:Patient3_Layer3}
  \includegraphics[width=0.4\textwidth]{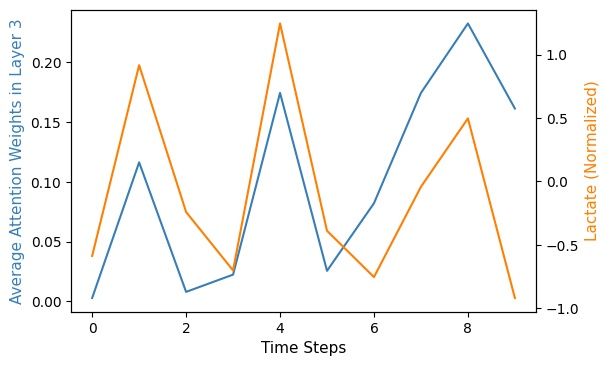}
}\qquad 
\subfigure[Delta SOFA score and attention weights (layer 4)]{%
  \label{fig:Patient3_Layer4}
  \includegraphics[width=0.4\textwidth]{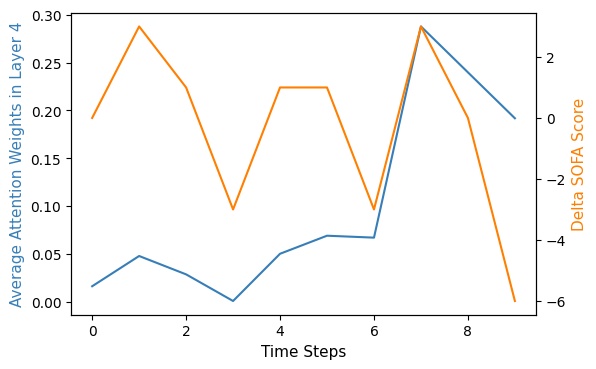}
}
}
\end{figure}

\end{document}